\documentclass[conference]{IEEEtran}
\IEEEoverridecommandlockouts
\usepackage{bm}
\usepackage{cite}
\usepackage{svg}
\usepackage{amsmath,amssymb,amsfonts}
\usepackage{graphicx}
\usepackage{textcomp}
\usepackage{xcolor}
\usepackage{caption}
\usepackage{subcaption}
\usepackage{algorithm}
\usepackage{algorithmic}
\usepackage{hyperref}
\def\BibTeX{{\rm B\kern-.05em{\sc i\kern-.025em b}\kern-.08em
    T\kern-.1667em\lower.7ex\hbox{E}\kern-.125emX}}

\newlength{\myfigwidth}
\ifCLASSOPTIONonecolumn
	\AtBeginDocument{\setlength{\myfigwidth}{0.8\linewidth}}
\else
	\AtBeginDocument{\setlength{\myfigwidth}{\linewidth}}
\fi
\newcommand{\myunit}[1]{%
	\ifmmode
		\mathrm{#1}
	\else
		$ \mathrm{#1} $
	\fi}
\newcounter{MYalgorithmic}

{%
	\vspace*{3pt}
	\hrule height 1pt}

\newcommand{\MYlabel}[1]{\def\@currentlabel{\theALG@line}\label{#1}}

\definecolor{myblue}{rgb}{0,0.4980,1} 
\definecolor{myred}{rgb}{0.8706,0.1608,0.0627} 

\begin{document}

\title{Towards Infant Sleep-Optimized Driving: Synergizing Wearable and Vehicle Sensing in Intelligent Cruise Control\\
}

\author{
Ruitao Chen,
Mozhang Guo, and Jinge Li \\
E-mail: rchen328@uwo.ca, mguo224@uwo.ca, and lij269@mcmaster.ca
}

\maketitle

\begin{abstract}
Automated driving (AD) has substantially improved vehicle safety and driving comfort, but their impact on passenger well-being, particularly infant sleep, is not sufficiently studied. Sudden acceleration, abrupt braking, and sharp maneuvers can disrupt infant sleep, compromising both passenger comfort and parental convenience. 
To solve this problem, this paper explores the integration of reinforcement learning (RL) within AD to personalize driving behavior and optimally balance occupant comfort and travel efficiency. 
In particular, we propose an intelligent cruise control framework that adapts to varying driving conditions to enhance infant sleep quality by effectively synergizing wearable sensing and vehicle data. Long short-term memory (LSTM) and transformer-based neural networks are integrated with RL to model the relationship between driving behavior and infant sleep quality under diverse traffic and road conditions. 
Based on the sleep quality indicators from the wearable sensors, driving action data from vehicle controllers, and map data from map applications, the model dynamically computes the optimal driving aggressiveness level, which is subsequently translated into specific AD control strategies, e.g., the magnitude and frequency of acceleration, lane change, and overtaking. Simulation experiments conducted in the CARLA environment indicate that the proposed solution significantly improves infant sleep quality compared to baseline methods, while preserving desirable travel efficiency.

\end{abstract}


\section{Introduction}
Vehicle occupant comfort has become an increasingly important focus in intelligent automotive applications\cite{domova2024comfort, zhang2024mitigation}, including smart cabins\cite{Chen2024Cockpits} and adaptive seating systems\cite{Zhan2025Seat}, aimed at enhancing the driving experience by reducing cognitive load and improving both physical and mental well-being. Despite these advancements, the specific comfort needs of a critical yet frequently overlooked passenger group—infants—remain underexplored in both research and industry. Infants require up to 18 hours of sleep per day\cite{butler2024association}, with high-quality sleep playing a vital role in neural development and physical growth\cite{liu2024childhood}. However, the in-vehicle environment presents considerable challenges, such as vibrations and noise, which can adversely affect infant sleep quality\cite{zhang2024mitigation}. Therefore, there is a pressing need for an infant-focused sleep enhancement solution to ensure a more comfortable travel experience and support optimal infant health.

A critical challenge in ensuring occupant comfort, particularly for infant sleep, is to accurately model the occupant's physiological and behavioural states while developing adaptive control algorithms for effective intervention. This requires real-time monitoring and assessment of physical and physiological indicators, e.g., movement and respiration patterns, along with dynamic adjustments to vehicle driving behaviour to foster a sleep-conducive environment. Conventional approaches in \cite{rajesh2023comfort, scheidel2024deep, wadi2024mitigating, shi2024ideal} use deterministic modelling of the physical state of vehicle occupants to adjust driving control to alleviate occupant discomfort. However, these methods do not capture the individual variability and dynamic nature of occupant comfort, particularly in real-time driving conditions. To address the limitation, the work in \cite{su2021study} proposes to utilize wearable sensors combined with machine learning-based feature extraction to assess occupants' physical and physiological states. Similarly, the works in \cite{wang2024reinforcement, barka2024driving} suggest leveraging in-situ human feedback as an indicator for reinforcement learning-based autonomous driving control, enabling a more interactive approach to riding comfort optimization.

Nevertheless, the above studies \cite{rajesh2023comfort, scheidel2024deep, wadi2024mitigating, siddiqi2022motion, barka2024driving, xiang2022comfort} fail to account for the diverse needs of vehicle occupants and the ever-changing driving environment, both of which are essential for real-time adaptive control to alleviate discomfort. First, comfort preferences can vary widely among individuals. For instance, a control algorithm optimized to improve ride comfort or facilitate infant sleep on rough roads may be ineffective for other occupants with different sensitivities to driving disturbances. Furthermore, driving factors such as road surface quality, traffic congestion, and intersections can vary continuously throughout a trip, directly impacting the effectiveness of comfort-enhancing measures. Consequently, algorithms calibrated for generalized driving conditions often lack the adaptability to address the unique characteristics of specific scenarios, thereby limiting their efficacy in optimizing occupant comfort.

To address such challenges, we propose an intelligent control framework that integrates wearable and vehicle sensing to deliver personalized, context-aware interventions for enhancing infant sleep quality. Specifically, physiological data collected from wearable sensors is used to assess infant sleep quality, while contextual driving information-such as braking and acceleration patterns, vehicle speed, and route conditions-is obtained from onboard sensors and applications (e.g., GPS systems and digital maps). This multimodal data is then processed by a reinforcement learning (RL) model to dynamically adapt the driving aggressiveness level, which is a setting similar to the driving style modes commonly used in automated driving systems, e.g., Tesla’s Full Self-Driving\cite{tesla_modely_manual}. To incorporate this mechanism into cruise control systems, each aggressiveness level is mapped to a control profile—defined by parameters such as speed, spacing, jerk, and acceleration—which is then used by cruise control algorithms to optimize ride comfort and driving efficiency. 
The key contributions of this paper are as follows:
\begin{itemize}
    \item We propose a novel intelligent control framework that integrates wearable sensing with vehicle sensing to enhance infant sleep quality in time varying driving environments.  
    \item An RL-based model is developed that incorporates various feature extraction and fusion techniques to dynamically adjust driving aggressiveness of cruise control. 
    \item The proposed control framework is evaluated in a simulated driving environment, and its performance is validated against multiple benchmark methods. 
\end{itemize}

This paper is structured as follows: Section \ref{sec:problem_statement_RLframework} presents the problem statement and introduces the RL framework for wearable sensing-based intelligent cruise control. 
Section \ref{sec:sysmodelling} describes the RL model developed for infant sleep enhancement. 
Section \ref{sec:proposed_soln} details the network models and algorithms, followed by a discussion of experiment design and results in Section \ref{sec:simulation_results}.
Finally, Section \ref{sec:conclusion} provides the conclusion.

\section{Problem Statement and Reinforcement Learning Framework}\label{sec:problem_statement_RLframework}
\begin{figure}[!t]
    \centering
    \includegraphics[width=\columnwidth]{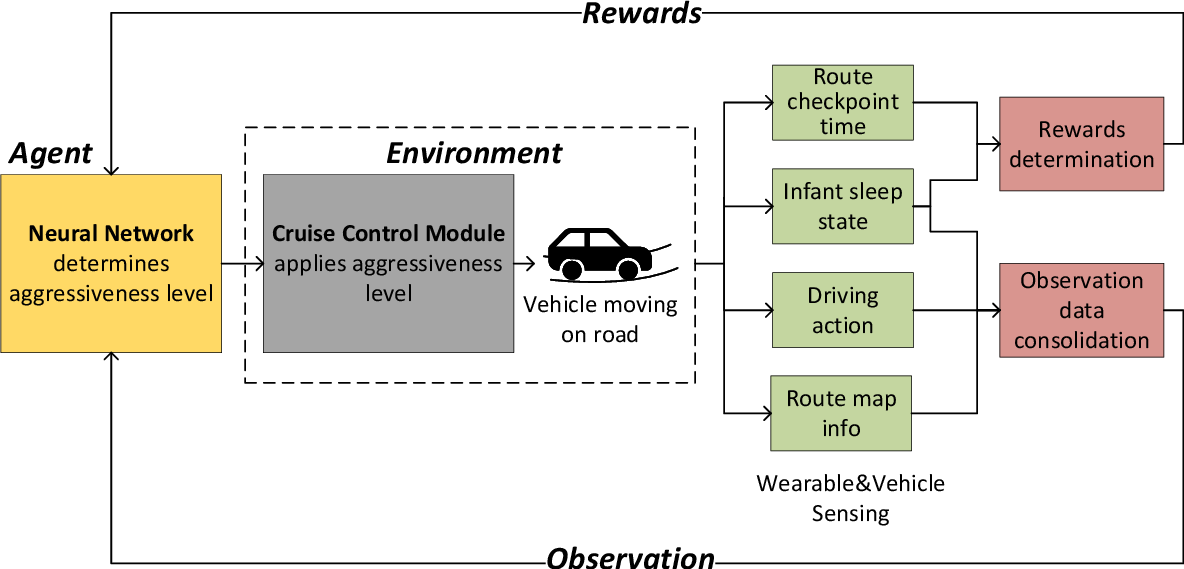}
    \caption{Reinforcement Learning Framework for Intelligent Cruise Control}
    \label{fig:RLStructure}
\end{figure}
\subsection{Problem Statement}
This paper focuses on addressing a specific occupant comfort problem—improving infant sleep quality—through optimal cruise control.
Similar to other occupant comfort issues, e.g., motion sickness, the sleep quality is primarily affected by vibrations and noise caused by vehicle motion. Intuitively, the negative impact on infant sleep can be alleviated by reducing driving aggressiveness, e.g., the magnitude of acceleration and braking, which can help limit the intensity and frequency of vehicle vibrations and noise. However, simply reducing driving aggressiveness is not an optimal solution, as it compromises traffic efficiency. For instance, reduced acceleration often leads to lower average speeds and, consequently, higher travel delays, especially on local roads where frequent acceleration and deceleration are required due to intersections, turns, and congestion. Therefore, an adaptive cruise control solution is required that can adjust driving aggressiveness to optimize infant sleep quality while also meeting users' travel time requirements.

\subsection{Reinforcement Learning Framework for Intelligent Cruise Control}
To improve the quality of infant sleep while satisfying travel time constraints, we propose a RL framework that intelligently determines the optimal driving aggressiveness level. This decision-making process leverages data from the occupant’s wearable devices, along with inputs from various onboard sensors and services. As shown in Fig.~\ref{fig:RLStructure}, the system architecture comprises the following components: a neural network module for action selection, a cruise control module that implements the selected actions within the vehicle’s control strategy, modules for monitoring occupant, vehicle, and road-related data, a reward calculation module, and a module for observation data consolidation.

Specifically, the neural network module determines its output, i.e., aggressiveness level, based on inputs including the infant’s sleep state, recent driving actions, and map information of the planned route. The cruise control module then translates this aggressiveness level into control parameters that influence acceleration, lane changes, and overtaking behavior. For instance, a higher aggressiveness level increases the upper bound on acceleration, reduces the minimum following distance to initiate a lane change, and lowers the speed threshold for triggering an overtaking maneuver. Consequently, the cruise control system becomes more active in executing lane changes and overtaking maneuvers, and applies higher acceleration when such actions are taken.

In addition, a set of modules continuously monitors the status of the occupant, vehicle, and road environment to support reward calculation and neural network decision-making. Specifically, the route checkpoint time tracking module evaluates whether the vehicle adheres to its schedule by comparing the actual time of arrival (ATA) with the estimated time of arrival (ETA) at each checkpoint along the planned route. The infant sleep state module tracks sleep state and quality using motion and physiological data collected from wearable sensors. Based on the collected information, rewards are calculated to achieve an optimal trade-off between the two objectives: improving sleep quality and optimizing trip efficiency. Similarly, driving action data (e.g., operations on the accelerator, brake, and steering wheel) and route map data (e.g., turns, intersections, speed limits, and lane counts) are recorded by two dedicated modules. Finally, data from the infant sleep, driving action, and route map modules are consolidated into an observation sample, which is fed into the neural network for decision-making.

\section{System modelling for aggressiveness control via reinforcement learning}\label{sec:sysmodelling}

\subsection{Overview}
\begin{figure}[!t]
    \centering
    \includegraphics[width=\columnwidth]{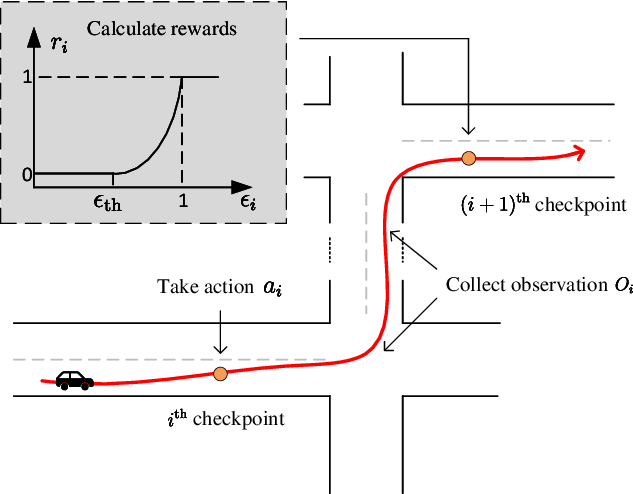}
    \caption{An illustration of the decision-making checkpoints along the route.}
    \label{fig:checkpoints}
\end{figure}

To effectively model the driving environment of the proposed RL algorithm, we first set multiple checkpoints on the planed route for decision-making. At each checkpoint, an action would be determined and taken by the RL agent. The corresponding rewards would be calculated and new observation data would be collected during the section between the current checkpoint and the next checkpoint. For instance, if the route is 100km, we can set 100 checkpoints that are evenly distributed over the entire route so that the vehicle passes a checkpoint every kilometer. As shown in Fig.~\ref{fig:checkpoints}, an action $a_i$ would be determined and taken when the vehicle reaches the $i^{th}$ checkpoint. Then observation data $O_i$ would be updated during the section between the $i^{th}$ checkpoint and the $(i+1)^{th}$ checkpoint. Finally, the reward $r_i$ for taking the action $a_i$ is calculated when the vehicle arrives at the $(i+1)^{th}$ checkpoint. Meanwhile, a new action would be determined for next route section based on the current observation $O_i$. The action, state and reward of all checkpoints along the route constitute a data sample of trip sequence. 
\subsection{Action Space}
The action determined by the neural network is used as aggressiveness level, denoted as $a_{i}$
 for the action determined at the $i^{th}$ checkpoint. 
 The aggressiveness level is designed as a discrete value in our work, and the set of all possible aggressiveness levels forms the action space, denoted as following:
 \begin{equation}
      \mathcal{A} = \{a^{\min}, a^{\min} + 1, ..., a^{\max}\},\label{eq:action}
 \end{equation}
where non-negative integers $a^{\min}$ and $a^{\max}$ 
 are the minimum and maximum possible value. The aggressiveness level is used by cruise control module when determining if it is needed to trigger lane change or overtaking actions, as well as the specific acceleration values. Note that there is no constraint on the continuity or discreteness of aggressiveness level, so the approach can also be adapted to control system with continuous action space, i.e., continuous value of aggressiveness level. 

\subsection{Cruise Control Environment}
The reinforcement learning environment includes a cruise control system and a vehicle running on the road. The cruise control system translates the aggressiveness level to specific control parameters, such as the magnitude of acceleration, the minimum distance between vehicles to enable lane change, and the speed threshold to trigger overtaking on a given lane. When the level of aggressiveness increases, these parameter values would lead the system to initiate lane changes and overtaking maneuvers more frequently, as well as to select faster acceleration under the same conditions, which could enhance travel efficiency while adversely affecting occupant comfort. Note that in this paper, we do not specify the types of parameters that are used to realize the desired aggressiveness level, with the aim of achieving a more general approach that is applicable across various control schemes. 

\subsection{Reward Function}
In our work, reward is designed to capture user’s requirement on travel time. Specifically, we compare the actual time of arrival (ATA) and the estimated time of arrival (ETA) at each checkpoint to assess how well the system adheres to the expected schedule. Let $t^E_i$ denote the ETA at the $i^{\text{th}}$ checkpoint, based on the route planning result, and $t^A_i$ denote the ATA, i.e., the actual elapsed time from the beginning of the trip to the $i^{\text{th}}$ checkpoint. The reward for taking an action at the $i^{\text{th}}$ checkpoint is determined at the $(i + 1)^{\text{th}}$ checkpoint, and is defined by
 \begin{equation}
     r_{i} = \mathcal{R}(\epsilon_i), \label{eq:rewards}
 \end{equation}
 where $\mathcal{R}(\cdot)$ denotes the reward function, and $\epsilon_i = t^E_{i}/t^{A}_i$ is the ratio between ETA and ATA.
 The $\mathcal{R}(\cdot)$ used by our work is shown in Fig.~2. In particular, when ATA is significantly higher than ETA, i.e., when $\epsilon_i$ is lower than a threshold $\epsilon_{th}$, the minimum reward is selected, i.e., $r_i=0$, because the vehicle is behind the schedule considerably. When $\epsilon_i > \epsilon_{th}$, the reward increases gradually with $\epsilon_{i}$, so a positive reward is assigned based on the extent of the delay. The reward would reach the maximum value, i.e., $r_i = 1$, when ETA is greater than ATA, i.e.,  when $\epsilon_i > 1$, because it indicates the vehicle is already ahead of its schedule. Note that we use an exponential function for the range $\epsilon_{th} < \epsilon_i < 1$, but other functions can be considered in the similar way. 
\subsection{Observation Space}
Observation space is modeled to include features related to the sleep state, driving action, and route map. To achieve a generic approach that is applicable to systems with different sensing capabilities, we select parameters that can be collected from common wearable sensors, vehicle control systems, and map applications. Then the value of these feature parameters collected during the section between the $i^{th}$ and $(i+1)^{th}$ checkpoints are used to determine the state vector $V_i$.
In particular, the parameters of infant sleep include the maximum motion $M_i^{\max}$ and average motion $M_i^{\text{avg}}$, which represent the maximum value and average value of the output of motion sensor in wearable devices, e.g.,  inertial measurement unit (IMU) in a smart wristband. In addition, the parameters of driving operation consist of the number of acceleration cycles $N_i^{\text{acc}}$, the number of turning signals 
$N_i^{\text{trn}}$ , maximum steering wheel angles $W_i$, and average vehicle speed $S_i^{\text{avg}}$. Here an acceleration cycle is defined as the occurrence of an acceleration command initiated once after a braking action. Finally, the features of route map include speed limit $S_i^{\text{lmt}}$ and the number of intersection crossings for left turn, right turn, and straight proceeding, denoted as 
$N_i^{\text{lt}}$, $N_i^{\text{rt}}$, $N_i^{\text{st}}$, respectively. Given that route map is pre-planned and the information is known in advance, the corresponding parameters for the next route section, i.e., the section between the $(i+1)^{th}$ and $(i+2)^{th}$ checkpoint, are also included in features, represented by $S_{i+1}^{\text{lmt}}$, $N_{i+1}^{\text{lt}}$, $N_{i+1}^{\text{rt}}$, and $N_{i+1}^{\text{st}}$ accordingly. Therefore, a state vector is defined as 
\begin{multline}
    V_i = [M_i^{\max}, M_i^{\text{avg}}, N_i^{\text{acc}}, N_i^{\text{trn}}, W_i, S_i^{\text{avg}}, S_i^{\text{lmt}}, \\N_i^{\text{lt}}, N_i^{\text{rt}}, N_i^{\text{st}}, S_{i+1}^{\text{lmt}}, N_{i+1}^{\text{lt}}, N_{i+1}^{\text{rt}}, N_{i+1}^{\text{st}}]^T. \label{eq:state}
\end{multline}
A single state vector cannot effectively discern the temporal patterns or trends of features. Inclusion of the information allows the agent to react to the feature dynamics more efficiently and make more responsive decision. Therefore, we designed an observation matrix that consists of the state from the $K$ most recent checkpoints, written as:
 \begin{equation}
     O_i = [V_{i-K + 1}, V_{i-K + 2}, ..., V_i]. \label{eq:observation}
 \end{equation}

 \subsection{Termination Condition of Each Episode}
 Since the objective of the proposed approach is to improve infant sleep quality, each training episode should terminate when the baby wakes up. Multiple algorithms\cite{imtiaz2021systematic} have been developed to precisely estimate sleep state based on various sensors input. In this paper, we estimate sleep state using motion sensors, e.g., IMU, as they are widely available in wearable devices. But more sophisticated algorithms using additional inputs, e.g., heart rate and respiratory rate, can be considered similarly. In particular, a large motion value detected by the motion sensor can indicate that the user has awakened, since the motion value is usually at a minimum level when the user is asleep. Therefore, the episode ends when the maximum motion $M_i^{\max}$ exceeds a threshold $M^{\text{thres}}$, i.e., when $M_i^{\max} \geq M^{\text{thres}}$.
 
\section{Network modules and algorithms}\label{sec:proposed_soln}
 
\begin{figure}[!t]
    \centering
    \includegraphics[width=\columnwidth]{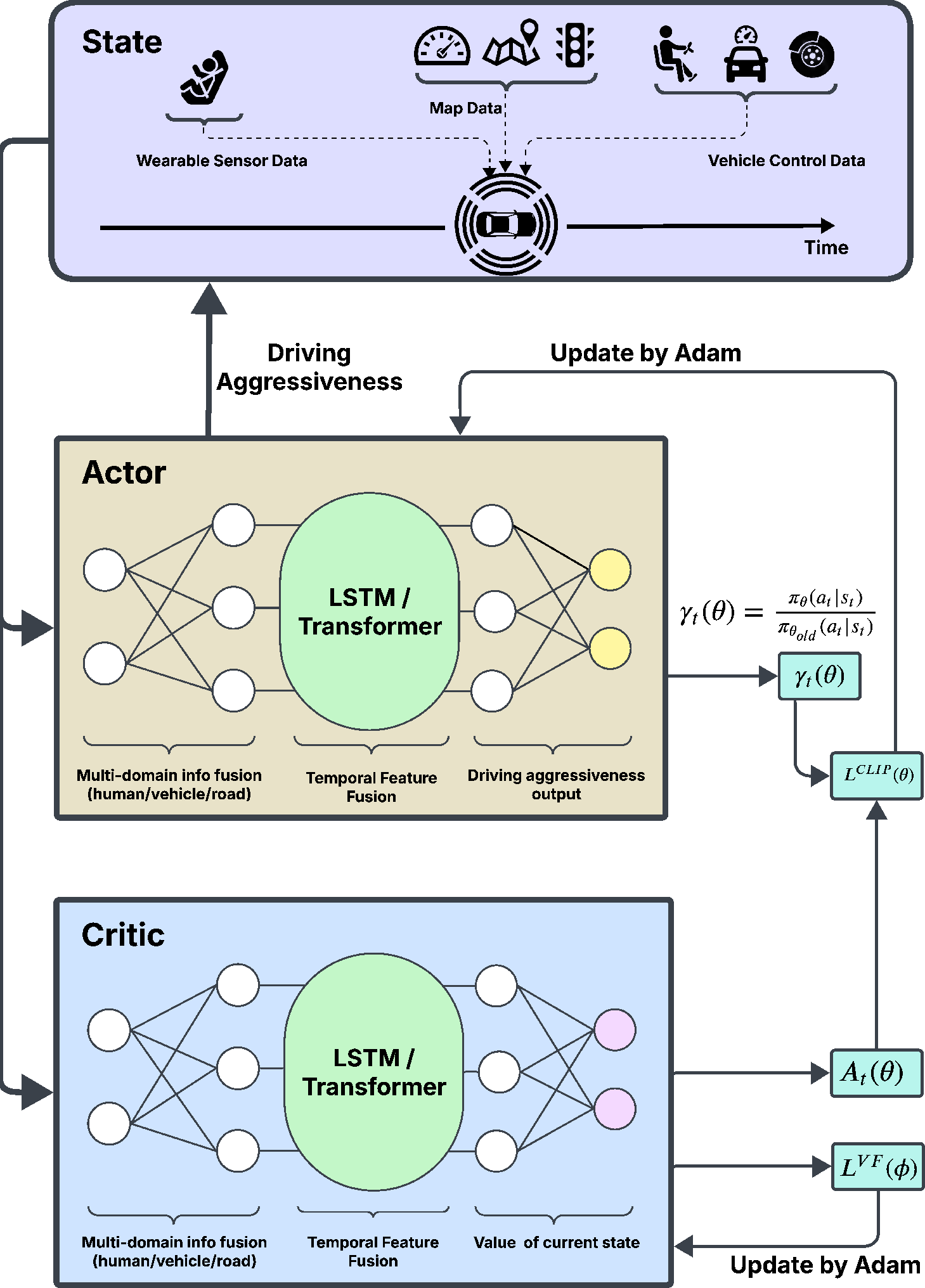}
    \caption{Proposed LSTM and Transformer based network structures integrated with PPO algorithm.}
    \label{fig:PPO}
\end{figure}

The Proximal Policy Optimization (PPO) algorithm \cite{schulman2017proximal} is adopted to train the proposed models. Given that the state vector encapsulates multi-dimensional temporal aspects of the driving environment, we propose a hybrid network architecture for effective processing. as shown in Fig.~\ref{fig:PPO}.

To achieve multi-dimensional information fusion and feature extraction, we employ a Multilayer Perceptron (MLP) network. For temporal feature learning, we investigate two deep neural network architectures: Long Short-Term Memory (LSTM) and Transformer models, as illustrated in Fig.~\ref{fig:PPO}. These architectures are designed to effectively capture sequential dependencies and long-range temporal correlations inherent in diverse driving scenarios. 

\textit{LSTM-based Structure with MLP:} This model integrates an LSTM network for sequential processing, coupled with MLP to extract multi-dimensional information, allowing the agent to learn temporal dependencies effectively. First, we use MLP layers to integrate information in the state vector, written as:
\begin{equation}
    \bm{V}_i^{l} = \text{ReLU}( \bm{W}_p\bm{V}_i^{l-1} + \bm{b}_p), W_p \in \mathbb{R}^{d \times h}, \label{eq:MLP_proc}
\end{equation}
where $\bm{V}_i^{l}$ is output of the $l^{\text{th}}$ layer of MLP and $\bm{W}_p$, $\bm{b}_p$ are the corresponding weight and bias parameters. Then, the outputs $\bm{V}^{L}_{i - K + t}$, for $t = 1, 2, ..., K$, from the last MLP layer is fed into the LSTM module, given by
\begin{equation}
    \bm{c}_t, \bm{h}_t = \text{LSTM}(\bm{c}_{t-1}, \bm{h}_{t-1}).
\end{equation}
The final hidden state $\bm{c}_t$ derived from the last LSTM output is used to calculate final result of the network:
\begin{equation}
    \bm{R}_t = \bm{W}_o\bm{h}_K + \bm{b}_o,
\end{equation}
where $\bm{W}_o$ and $\bm{b}_o$ are parameters of the output layer.

\textit{Transformer-based Structure with MLP:} This model leverages a Transformer network to capture long-range dependencies in the temporal domain, while the MLP processes multi-dimensional information, enhancing the model's ability to generalize across diverse driving conditions. Specifically, 
each state vector is first processed by the MLP layers as in \eqref{eq:MLP_proc}, producing a temporal sequence  $\bm{\hat{V}} = \{\bm{V}^{L}_{i -K + 1},\bm{V}^{L}_{i -K + 2} , \dots, \bm{V}^{L}_i\}$. To incorporate global temporal context, a learnable classification token $v_{\text{class}}$ is prepended, forming the augmented input to the Transformer encoder:
\begin{equation}
\bm{V}_{t} = [v_{\text{class}}, \bm{\hat{V}}] = [v_{\text{class}}, \bm{V}^{L}_{i - K + 1}, \dots, \bm{V}^{L}_i].
\end{equation}

This modified sequence is then passed through the transformer encoder, where self-attention is computed as follows:
\begin{equation}
    \bm{Z}_t = \text{softmax}\left(\frac{\bm{Q}_C \bm{K}_C^T}{\sqrt{d}}\right) \bm{V}_C,
\end{equation}
where $\bm{Q}_C = \bm{V}_t \bm{W}_Q$, $\bm{K}_C = \bm{V}_t \bm{W}_K$, and $\bm{V}_C = \bm{V}_t \bm{W}_V$,
and they represent the query, key, and value matrices derived from the input $\bm{V}_t$ using the learnable  weights $\bm{W}_Q$, $\bm{W}_K$, and $\bm{W}_V$, respectively. The attention output $\bm{Z}_t$ encodes temporal interactions within the sequence. 
The final output representation $\bm{O}_i^{final}$ of the environment state is the first token in $\bm{Z}_t$.

\textit{PPO Traning:} The PPO-based training process is introduced in Algorithm \ref{alg:algo1}. In Step 3, the agent first collects driving information consisting of states \eqref{eq:state}, actions \eqref{eq:action}, and rewards \eqref{eq:rewards} through interactions with various driving environments.
This information serves as the foundation for computing the policy loss and value loss, which are utilized to update the weights of the actors and critic networks.

The policy loss quantifies the discrepancy between the current policy and the updated policy derived from the collected trajectories. It encourages the agent to increase the probability of actions that lead to higher rewards while decreasing the probability of actions that result in lower rewards. Let $\theta$ denote the parameters of the actor network and $\gamma_t(\theta))$ denote the ratio of current and old policy output, i.e $\gamma_t(\theta) = \frac{\pi_\theta (a_t | s_t)}{\pi_{\theta_{\text{old}}} (a_t | s_t)}$. The policy loss is formulated as:
\begin{equation}
    \mathcal{L}^{\text{CLIP}}(\theta) = \mathbb{E} \left[\min \left(\gamma_t(\theta) A_t(\theta), 
    g(\epsilon,\gamma_t(\theta))A_t(\theta) \right) \right],
    \label{eq:policyloss}
\end{equation}
where $g(\epsilon,\gamma_t(\theta)) = \text{clip}(\gamma_t(\theta), 1 - \epsilon, 1 + \epsilon)$ calculates clipped surrogate objective\cite{schulman2017proximal}. The value loss measures the discrepancy between the estimated value function and the actual rewards obtained during interactions. By minimizing this loss, the value function is refined to better approximate the expected cumulative rewards. The value loss is given by:
\begin{equation} 
\mathcal{L}^{VF}(\phi) = \mathbb{E} \left[\left( V_{\phi}(s_t) - R_t \right)^2 \right]. \label{eq:valueloss}
\end{equation}
Here $\phi$ represents the parameters of the critic network, $V_{\phi}(s_t)$ is the estimated value function under policy $\pi_{\theta}$, and $R_t$ denotes the reward-to-go, which is the cumulative discounted reward from time step t onward. Finally, the equations \eqref{eq:policyloss} and \eqref{eq:valueloss} are used to update the policy and value network parameters as described in Step 6-7 in Algorithm \ref{alg:algo1}.
\begin{algorithm}
\caption{PPO training for aggressiveness level determination networks}\label{alg:algo1}
\begin{algorithmic}[1]

\STATE Initial policy parameter $\theta$, initial value function parameter $\phi$
\FOR{k = 0, 1, 2, ...}
   \STATE Collect set of trip sequences $D_k = {\tau_i}$ by running policy $\pi_k = \pi(\theta_k)$ in the environment
   \STATE Compute rewards-to-go ${R}_t$
   \STATE Compute advantage estimates $A_t(\theta)$
   \STATE Update policy (used for aggressiveness level calculation) by minimizing the policy loss in equation \eqref{eq:policyloss}
   \STATE Fit value function by regression on mean-squared error in equation \eqref{eq:valueloss}
\ENDFOR
\end{algorithmic}
\end{algorithm}

\section{Simulation}\label{sec:simulation_results}

\subsection{Simulation Setup}
Experiments are conducted using the CARLA\cite{dosovitskiy2017carla} simulator (v0.9.13) in Town07, a mixed urban environment. The simulated vehicle is controlled by CARLA's TrafficManager Autopilot, which follows given routes while obeying traffic rules. The PPO model receives input from a simulated 3-axis wrist-mounted IMU sampled at 100Hz, as well as recent vehicle control signals, e.g., steering, turning signals, acceleration, brake, velocity, and local route context from CARLA’s internal map. Since CARLA is not capable of simulating human body dynamics, wrist acceleration $a_{\text{wrist}}(t)$ is computed using a second-order mass-spring-damper model to approximate compliant transmission through the seat and infant’s arm~\cite{mizrahi2015}:
\begin{equation}
\alpha_{\text{wrist}}(t) = \alpha_{\text{car}}(t) + \frac{1}{m} \left( -k x(t) - c v(t) \right),
\end{equation}
where $\alpha_{\text{car}}$ is the vehicle acceleration retrived from onboard IMU and $v(t)$ is the vehicle speed. The configuration parameters are set as
$m = 0.4\text{kg}$, $k = 50\text{N/m}$, and $c = 1.0\text{Ns/m}$. The IMU magnitude is monitored over a 3-second (300-sample) window, and the infant is labeled awake if $\max \lVert \alpha_{\text{wrist}} \rVert > 2\,\text{m/s}^2$.  Aggressiveness level $a_i$ is selected from the space $\mathcal{A} = \{0, 1, ..., 10\}$ and is implemented in Autopilot through configuration parameters, including Speed difference percentage $P_s$,  Following distance $P_\text{d}$ (m), Path update interval $P_\text{u}$ (ms), Auto Lane change enable $P_\text{a}$, and Keep right percentage $P_\text{r}$. These parameters are computed as follows: $P_\text{s}= 1.5 \times a_i$, $P_\text{d}= 6.0 - 0.4 \times a_i$, $P_u= 1000 - 70 \times a_i$, $P_\text{a}= \mathbb{I}\{a_i \geq 2\}$, and $P_r= 100 - 10 \times a_i$.

The PPO model outputs a continuous-valued driving aggressiveness level and is trained over 2000 iterations with 20 steps per update, using 100 different traffic condition settings and running routes in Town07. The Adam optimizer is used with a learning rate of $1 \times 10^{-4}$, $\gamma = 0.99$. Two policy architectures are evaluated: (1) an LSTM-based policy with sequence length 5, hidden size 128, 2 layers, and 0.1 dropout; and (2) a Transformer-based policy with 2 encoder layers, 4 attention heads, 128-dimensional embeddings, and learned positional encoding. Training the RL agent takes approximately 30 minutes on a single NVIDIA RTX 3060 GPU.

\subsection{Result Discussion}
\begin{figure}[!t]
    \centering
    \includegraphics[width=\columnwidth]{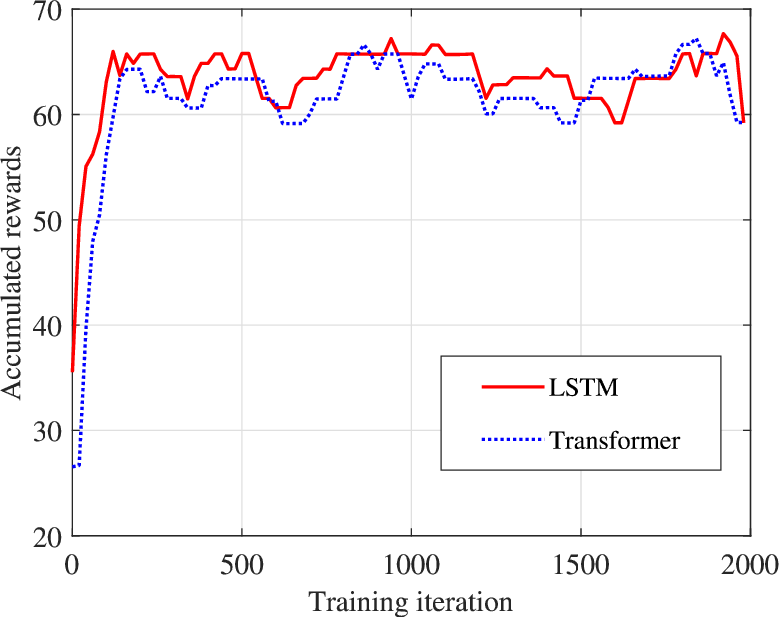}
    \caption{Convergence process of the proposed algorithm for LSTM and Transformer based network structures.}
    \label{fig:convergence}
\end{figure}
First, we evaluate the convergence performance of the PPO algorithm using two proposed network models. Fig.~\ref{fig:convergence} shows the accumulated rewards over training iterations. Both models converge within 500 iterations, corresponding to 10,000 training steps, as each iteration updates model parameters 20 times using a sequence sampled during that iteration. Note that an individual model is trained for each environment setting (i.e., vehicle and occupant), so the available data samples—i.e., trip sequences—are limited in the real world. We use a dataset of 100 trip sequences for the tested setting. Each iteration randomly samples one sequence, resulting in an average of five training iterations per sequence before convergence.

\begin{figure}[!t]
    \centering
    \includegraphics[width=\columnwidth]{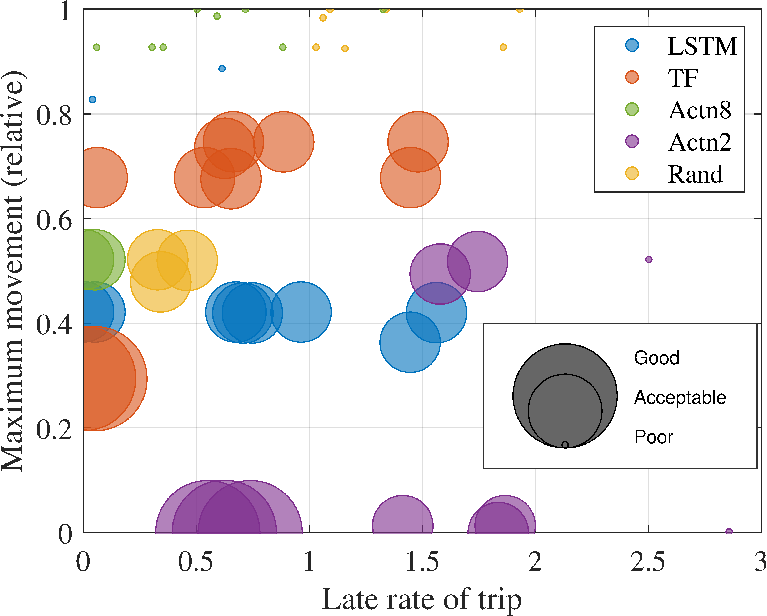}
    \caption{Relative maximum movement of occupant vs late rate of trip with different algorithms (both axes shown as ratios).}
    \label{fig:cloud}
\end{figure}

Next we investigate the proposed approach in terms of sleep quality improvement and trip delay. Sleep quality is quantified by the relative maximum movement of the occupant, defined as the normalized peak value of the motion sensor (i.e., IMU) output during a trip. Trip delay is measured by the trip late rate, calculated as $\max \left( {\left( t_{I}^{A}-t_{I}^{E} \right)}/{t_{I}^{E},0}\; \right)$, where $I$ denotes the final checkpoint of the trip, $t_{I}^{E}$ the ETA, and $t_{I}^{A}$ the ATA. As shown in Fig.~\ref{fig:cloud}, we calculate these metrics across 10 randomly sampled trip sequences for the proposed LSTM and Transformer (TF) algorithms, as well as three baselines: Actn8, Actn2, and Rand. In Actn8 and Actn2, a fixed aggressiveness level of 8 and 2, respectively, is used at each step. In Rand, the aggressiveness level is randomly sampled from $\mathcal{A}$.
To visualize performance differences, samples are categorized based on metric thresholds. A sample is labeled \textit{good} (large bubble) if both metrics are low, i.e., relative maximum movement $<0.4$ and trip late rate $<1$. It is labeled \textit{poor} (small bubble) if either metric is high, i.e., relative maximum movement $>0.8$ or trip late rate $>2$. All other samples are classified as \textit{acceptable} (medium bubble).  

It can be seen that LSTM achieves the optimal trade-off between sleep quality and trip latency across most trip samples. While TF performs similarly in latency, it shows higher maximum movement in several cases (e.g., more samples with maximum movement $>0.6$), indicating lower sleep quality. Both proposed models outperform benchmarks: Actn8 shows more samples with maximum movement $>0.9$, and Actn2 exhibits more samples with late rate $>1.5$. These results confirm the design assumption that increased aggressiveness reduces latency but degrades sleep quality. The Rand, with randomly assigned aggressiveness, shows no consistent performance pattern.

\begin{figure}[!t]
    \centering
    \includegraphics[width=\columnwidth]{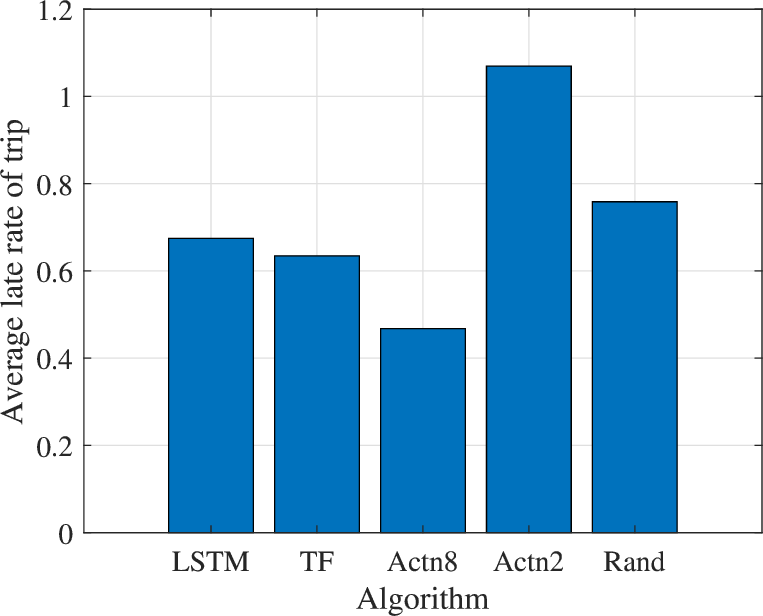}
    \caption{Average late rate with different algorithms.}
    \label{fig:late_ratio}
\end{figure}

\begin{figure}[!t]
    \centering
    \includegraphics[width=\columnwidth]{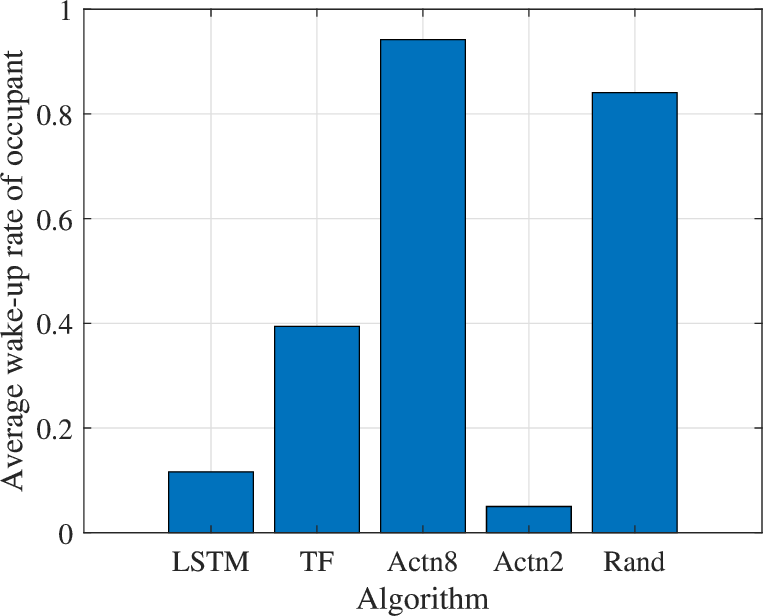}
    \caption{Average wake-up rate with different algorithms.}
    \label{fig:wake_up_ratio}
\end{figure}

We also evaluate the average performance across multiple samples, as shown in Fig.~\ref{fig:late_ratio} and Fig.~\ref{fig:wake_up_ratio}. Fig.~\ref{fig:late_ratio} reports the average trip late rate across all trip sequences, while Fig.~\ref{fig:wake_up_ratio} shows the percentage of sequences terminated due to occupant wake-up. Consistent with Fig.~\ref{fig:cloud}, the proposed LSTM and TF models achieve a better trade-off between latency and sleep quality than the benchmark methods. Although LSTM and TF perform similarly in average late rate, LSTM yields a lower wake-up rate, indicating better sleep quality. Notably, both excessively high and low aggressiveness lead to undesirable outcomes—e.g., Actn8 exhibits the highest wake-up rate, and Actn2 the highest late rate. The performance of Rand lies between those of Actn2 and Actn8 across both metrics.

\section{Conclusions}\label{sec:conclusion}
This work presents an intelligent cruise control framework that combines wearable sensing and vehicle data to enhance occupant comfort, with a focus on infant sleep under varying driving conditions. The system leverages reinforcement learning with novel network modules to model the impact of driving behavior on sleep quality and travel efficiency. It generates driving aggressiveness decisions based on multimodal inputs, including sleep-related signals from wearables, vehicle control data, and route information. Simulations show that the approach adapts effectively to dynamic driving conditions and outperforms baselines in improving sleep quality without sacrificing trip efficiency. Future work will address practical deployment challenges such as sensor latency and system integration, and explore additional physiological signals for more accurate sleep state estimation.

\bibliographystyle{IEEEtran}
\bibliography{IEEEabrv,Ref}

\begin{thebibliography}{10}
\providecommand{\url}[1]{#1}
\csname url@samestyle\endcsname
\providecommand{\newblock}{\relax}
\providecommand{\bibinfo}[2]{#2}
\providecommand{\BIBentrySTDinterwordspacing}{\spaceskip=0pt\relax}
\providecommand{\BIBentryALTinterwordstretchfactor}{4}
\providecommand{\BIBentryALTinterwordspacing}{\spaceskip=\fontdimen2\font plus
\BIBentryALTinterwordstretchfactor\fontdimen3\font minus \fontdimen4\font\relax}
\providecommand{\BIBforeignlanguage}[2]{{%
\expandafter\ifx\csname l@#1\endcsname\relax
\typeout{** WARNING: IEEEtran.bst: No hyphenation pattern has been}%
\typeout{** loaded for the language `#1'. Using the pattern for}%
\typeout{** the default language instead.}%
\else
\language=\csname l@#1\endcsname
\fi
#2}}
\providecommand{\BIBdecl}{\relax}
\BIBdecl

\bibitem{domova2024comfort}
V.~Domova, R.~M. Currano, and D.~Sirkin, ``Comfort in automated driving: A literature survey and a high-level integrative framework,'' \emph{Proceedings of the ACM on Interactive, Mobile, Wearable and Ubiquitous Technologies}, vol.~8, no.~3, pp. 1--23, Sep. 2024.

\bibitem{zhang2024mitigation}
Y.~Zhang, H.~Zhao, C.~Hu, Y.~Tian, Y.~Li, X.~Jiao, and G.~Wen, ``Mitigation of motion sickness and optimization of motion comfort in autonomous vehicles: Systematic survey,'' \emph{IEEE Transactions on Intelligent Transportation Systems}, vol.~25, no.~12, pp. 21\,737--21\,756, Oct. 2024.

\bibitem{Chen2024Cockpits}
H.~Chen, R.~Gao, L.~Fan, E.~Liu, W.~Li, R.~Tan, Y.~Li, L.~He, and D.~Cao, ``Scenario-function system for automotive intelligent cockpits: Framework, research progress and perspectives,'' \emph{IEEE Transactions on Intelligent Vehicles}, vol.~9, no.~5, pp. 4890--4904, Mar. 2024.

\bibitem{Zhan2025Seat}
H.~Zhan, P.~Liu, X.~Xia, D.~Ning, and H.~Du, ``Modeling and vibration control of a three-degree-of-freedom electrically interconnected seat suspension system for heavy-duty vehicle,'' \emph{IEEE Transactions on Industrial Electronics}, pp. 1--10, 2025.

\bibitem{butler2024association}
B.~Butler, R.~Burdayron, G.~Mazor-Goder, C.~Lewis, M.~Vendette, B.~Khoury, and M.-H. Pennestri, ``The association between infant sleep, cognitive, and psychomotor development: a systematic review,'' \emph{Sleep}, vol.~47, no.~11, p. zsae174, Nov. 2024.

\bibitem{liu2024childhood}
J.~Liu, X.~Ji, S.~Pitt, G.~Wang, E.~Rovit, T.~Lipman, and F.~Jiang, ``Childhood sleep: physical, cognitive, and behavioral consequences and implications,'' \emph{World Journal of Pediatrics}, vol.~20, no.~2, pp. 122--132, Feb. 2024.

\bibitem{rajesh2023comfort}
N.~Rajesh, Y.~Zheng, and B.~Shyrokau, ``Comfort-oriented motion planning for automated vehicles using deep reinforcement learning,'' \emph{IEEE Open Journal of Intelligent Transportation Systems}, vol.~4, pp. 348--359, May 2023.

\bibitem{scheidel2024deep}
H.~Scheidel, H.~Asadi, T.~Bellmann, A.~Seefried, S.~Mohamed, and S.~Nahavandi, ``A deep reinforcement learning based motion cueing algorithm for vehicle driving simulation,'' \emph{IEEE Transactions on Vehicular Technology}, vol.~73, no.~7, pp. 9696--9705, Mar. 2024.

\bibitem{wadi2024mitigating}
A.~Wadi, M.~Abdel-Hafez, and M.~A. Jaradat, ``Mitigating motion sickness in autonomous vehicles for improved passenger comfort,'' \emph{IEEE Access}, vol.~12, pp. 62\,709--62\,718, Apr. 2024.

\bibitem{shi2024ideal}
Z.~Shi, L.~He, M.~Wang, Y.~Bian, S.~Cui, and P.~Chen, ``Ideal comfort ellipses and comfort dynamics model for mitigating motion sickness in battery electric vehicle,'' \emph{Vehicle System Dynamics}, pp. 1--17, Dec. 2024.

\bibitem{su2021study}
H.~Su and Y.~Jia, ``Study of human comfort in autonomous vehicles using wearable sensors,'' \emph{IEEE Transactions on Intelligent Transportation Systems}, vol.~23, no.~8, pp. 11\,490--11\,504, Aug. 2021.

\bibitem{wang2024reinforcement}
\BIBentryALTinterwordspacing
Y.~Wang, L.~Liu, M.~Wang, and X.~Xiong, ``Reinforcement learning from human feedback for lane changing of autonomous vehicles in mixed traffic,'' 2024. [Online]. Available: \url{https://arxiv.org/abs/2408.04447}
\BIBentrySTDinterwordspacing

\bibitem{barka2024driving}
R.~E. Barka and I.~Politis, ``Driving into the future: A scoping review of smartwatch use for real-time driver monitoring,'' \emph{Transportation Research Interdisciplinary Perspectives}, vol.~25, p. 101098, May 2024.

\bibitem{siddiqi2022motion}
M.~R. Siddiqi, S.~Milani, R.~N. Jazar, and H.~Marzbani, ``Motion sickness mitigating algorithms and control strategy for autonomous vehicles,'' \emph{IEEE transactions on intelligent transportation systems}, vol.~24, no.~1, pp. 304--315, Oct. 2022.

\bibitem{xiang2022comfort}
J.~Xiang and L.~Guo, ``Comfort improvement for autonomous vehicles using reinforcement learning with in-situ human feedback,'' \emph{SAE Technical Paper}, Mar. 2022.

\bibitem{tesla_modely_manual}
\BIBentryALTinterwordspacing
``Model {Y} owner's manual,'' 2025. [Online]. Available: \url{https://www.tesla.com/ownersmanual/modely/en_us/}
\BIBentrySTDinterwordspacing

\bibitem{imtiaz2021systematic}
S.~A. Imtiaz, ``A systematic review of sensing technologies for wearable sleep staging,'' \emph{Sensors}, vol.~21, no.~5, p. 1562, Feb. 2021.

\bibitem{schulman2017proximal}
\BIBentryALTinterwordspacing
J.~Schulman, F.~Wolski, P.~Dhariwal, A.~Radford, and O.~Klimov, ``Proximal policy optimization algorithms,'' 2017. [Online]. Available: \url{https://arxiv.org/abs/1707.06347}
\BIBentrySTDinterwordspacing

\bibitem{dosovitskiy2017carla}
A.~Dosovitskiy, G.~Ros, F.~Codevilla, A.~Lopez, and V.~Koltun, ``Carla: An open urban driving simulator,'' in \emph{Proceedings of the 1st Annual Conference on Robot Learning}, 2017, pp. 1--16.

\bibitem{mizrahi2015}
J.~Mizrahi, ``Mechanical impedance and its relations to motor control, limb dynamics, and motion biomechanics,'' \emph{Journal of Medical and Biological Engineering}, vol.~35, no.~1, pp. 1--20, Jan. 2015.

\end{thebibliography}
\end{document}